\documentclass[10pt,twocolumn,letterpaper]{article}

\usepackage{cvpr}

\usepackage{epsfig}
\usepackage{graphicx}
\usepackage{amsmath}
\usepackage{amssymb}
\usepackage{bm}
\usepackage{multirow}
\usepackage{multicol}
\usepackage[table,xcdraw]{xcolor}
\usepackage{float}
\usepackage{colortbl,hhline}
\usepackage{booktabs}
\usepackage{array,eucal}
\usepackage{appendix}

\usepackage{authblk}

\usepackage[margin=4pt,font=small,labelfont=bf,labelsep=period,tableposition=top]{caption}

\usepackage[pagebackref=true,breaklinks=true,letterpaper=true,colorlinks,bookmarks=false]{hyperref}

\cvprfinalcopy 

\begin{document}

\title{PGL: Prior-Guided Local Self-supervised Learning \\ for 3D Medical Image Segmentation\vspace{-0.4cm}}
\author[1,2]{\rm Yutong Xie\thanks{YX and JZ contributed equally. Work was done when YX and JZ were visiting The University of Adelaide.}}
\author[1,2]{\rm Jianpeng Zhang$^*$}
\author[1]{\rm Zehui Liao}
\author[1]{\rm Yong Xia}
\author[2]{\rm Chunhua Shen}
\affil[1]{~School of Computer Science and Engineering, Northwestern Polytechnical University, China}
\affil[2]{~The University of Adelaide, Australia}
\affil[ ]{{\tt\small \{xuyongxie, james.zhang, merrical\}@mail.nwpu.edu.cn}; {\tt\small yxia@nwpu.edu.cn}; {\tt\small chunhua.shen@adelaide.edu.au}}

\maketitle

\begin{abstract}
It has been widely recognized that the success of deep learning in image segmentation relies overwhelmingly on a myriad amount of densely annotated training data, which, however, are difficult to obtain due to the tremendous labor and expertise required, particularly for annotating 3D medical images.
Although self-supervised learning (SSL) has shown great potential to address this issue, most SSL approaches focus only on image-level global consistency, but ignore the local consistency which plays a pivotal role in capturing structural information for dense prediction tasks such as segmentation.
In this paper, we propose a Prior-Guided Local (PGL) self-supervised model that learns the region-wise local consistency in the latent feature space. 
Specifically, we use the spatial transformations, which produce different augmented views of the same image, as a prior to deduce the location relation between two views, which is then used to align the feature maps of the same local region but being extracted on two views.
Next, we construct a local consistency loss to minimize the voxel-wise discrepancy between the aligned feature maps.
Thus, our PGL model learns the distinctive representations of local regions, and hence is able to retain structural information. This ability is conducive to downstream segmentation tasks. 
We conducted an extensive evaluation on four public computerized tomography (CT) datasets that cover 11 kinds of major human organs and two tumors. The results indicate that using pre-trained PGL model to initialize a downstream network leads to a substantial performance improvement over both random initialization and the initialization with global consistency-based models.
Code and pre-trained weights will be made available at:   
\def\UrlFont{\rm\small\ttfamily}
\url{https://git.io/PGL}.
\end{abstract}

\section{Introduction}
Automated and accurate segmentation of 3D medical images acts a pivotal part in aiding medical professionals to make diagnoses, surgical planning, and prognosis. 
Currently, medical image segmentation approaches are overwhelmingly based on deep convolutional neural networks (DCNNs)~\cite{3dUnet,Unet,Unet++,ZhangTMI20,MedSeg_survey}. 
Despite their ameliorative accuracy, these approaches rely heavily on large amounts of densely annotated training data. Such a dataset is hard to create, since annotating 3D medical images at voxel level is labor-intensive and requires a high degree of expertise that is not always available~\cite{ZZW, CVPR20_smalldata, CVPR19_smalldata}.

\begin{figure}[!t]
	\begin{center}
		{\includegraphics[width=1.0\linewidth]{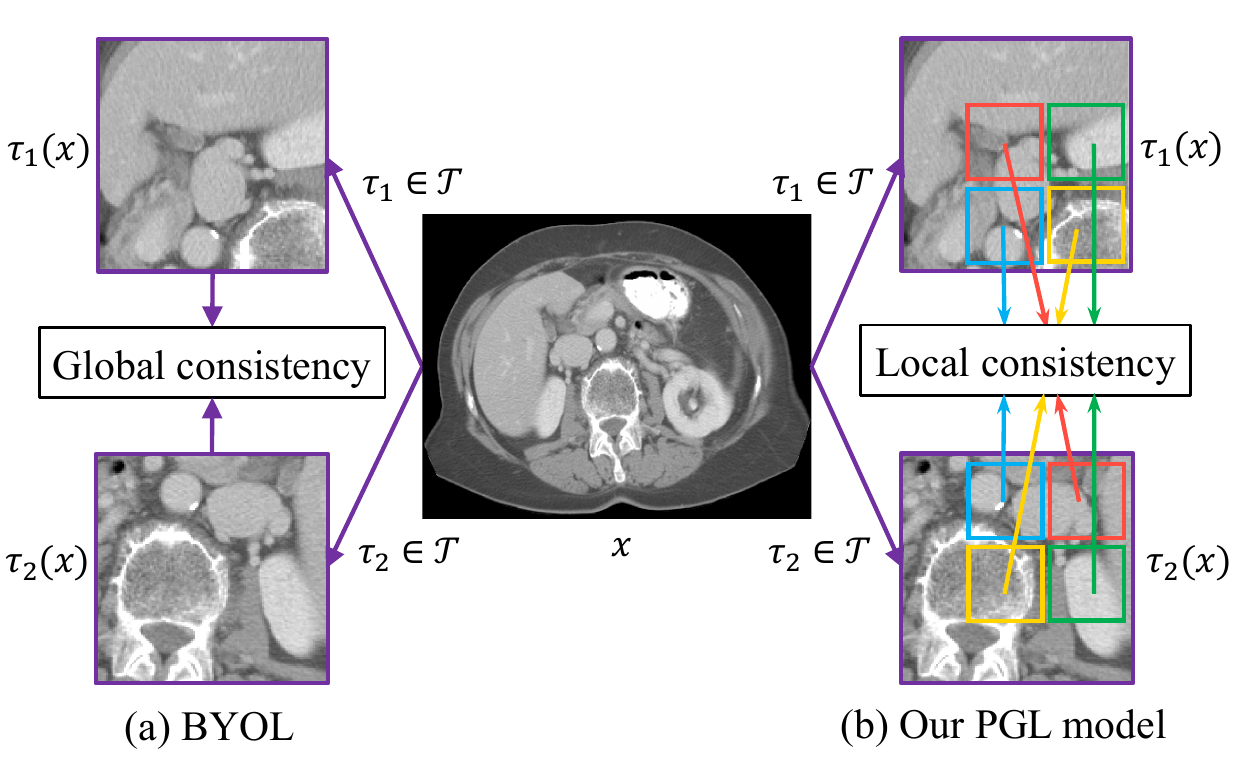}}
	\end{center}
 	\vspace{-0.5cm}
	\caption{
	Intuitive explanation of BYOL~\cite{BYOL} and our PGL model.
	Given two different augmented views of an image, BYOL focuses on learning the global consistency between a pair of views, 
	while our PGL employs the prior location relation between a pair of views to learn the local consistency between the features of the same region on different views. Here, the same color rectangles within $\tau_{1}(\bm{x})$ and $\tau_{2}(\bm{x})$ represent the same local region.
	Compared to BYOL, PGL can lean more structural information, which contributes to downstream medical image segmentation tasks. 
	The $\mathcal{T}$ is the family of data augmentations. 
	}	
	\label{fig:fig1}
	\vspace{-0.1cm}
\end{figure}

Self-supervised learning (SSL)~\cite{ZZW,CVPR20_smalldata,SSL_survey_PAMI,SSL_survey_arXiv} is able to learn visual representations without using human annotations, and has a proven track record in initializing DCNNs to perform a downstream task which has limited annotations.
Recent SSL methods are usually based on the contrastive learning~\cite{CPC,CMC,PIPL,MOCO,AMDIM,DIM}, which exploits unlabeled data according to the assumption that the representations of augmented views of the same image (positive pairs) should be consistent, and  the representations of augmented views of different images (negative pairs) should be inconsistent.
In contrast, the Bootstrap Your Own Latent (BYOL) model~\cite{BYOL} learns image representations using a dual online-target path paradigm that trains the online network on an augmented view of an image to predict the representation of another augmented view of the same image produced by the target network. Thus, BYOL sets negative pairs free, escapes the contrastive prototype, and achieves the state-of-the-art performance in SSL.

Despite its success, BYOL focuses only on learning the global consistency between each positive pair, but ignoring the local consistency which plays a pivotal role in capturing structural information for dense prediction tasks such as image segmentation.
Learning the local consistency, however, is not easy, since the transformation that generates a view of the input image alters voxel locations drastically. Fortunately, no matter which transformation is performed, the spatial location relation between two augmented views can be extrapolated as long as they are transformed from the same image. Such relation can serve as a prior to guide the network towards learning consistent local representations from unlabeled data. 

In this paper, we propose a Prior-Guided Local (PGL) self-supervised model for 3D medical image segmentation, which employs the prior location relation between a pair of positive images to learn the local consistency between the features of the same region.
This model is composed of a data augmentation module for image view generation and a prior dual-path module for feature extraction. 
The information about the transformation used for data augmentation (\textit{i.e.}, flipping, cropping, and scaling) is introduced to the dual online-target path paradigm as a prior, aiming to align the image features of the same region but extracted on different views. Then we construct a local consistency loss to minimize the discrepancy between each pair of aligned features.
In this way, our model can learn better representations of local regions, capture more structural information, and is more suitable for downstream segmentation tasks than those based on the global consistency.
Extensive experiments on four public datasets demonstrate the effectiveness of the proposed PGL model.
Our contributions are three-fold.
\begin{itemize}
\vspace{-0.2cm}
\item We propose the PGL self-supervised model that can learn effective visual representations without using any annotations, and thus alleviates the dependency on a large-scale densely-annotated training dataset. This is significant, since developing annotation-efficient deep models for medical image segmentation is a long-standing challenge.
\vspace{-0.2cm}
\item We use the spatial transformation of the input as a prior to draw consistent latent representations of the same image region on different views, and guarantee such consistency via minimizing a local consistency loss. Thus, the network is able to grasp more structural representations for medical image segmentation.
\vspace{-0.2cm}
\item We perform extensive experiments on four public CT datasets. The results show that our PGL model substantially outperforms both random initialization and BYOL, a global consistency-based SSL method.
	
\end{itemize}

\section{Related Work}
Despite of their success, the performance of DCNNs in medical image segmentation depends heavily on the scale of densely annotated training dataset.
Recently, SSL is widely recognized as a hopeful means to alleviate such dependency, since it can learn useful image representations from unlabeled data~\cite{ZZW, Medical_SSL0, Medical_SSL1, CVPR20_smalldata,SSL_survey_PAMI,SSL_survey_arXiv}.
Most SSL approaches can be summarized into two categories.

\noindent
\textbf{Generative SSL approaches}, including image generation with GAN~\cite{GAN}, super-resolution~\cite{Super_resolution}, image in-painting~\cite{Inpainting}, image decoupling~\cite{Decoupling}, and image colorization~\cite{Colorization}, learn image representations using training images themselves as the supervision to generate images pixel-for-pixel. The pixel-for-pixel generation, however, has a high computational cost and is hard to learn more abstract latent representations.

\noindent
\textbf{Discriminative SSL approaches} construct discriminative pretext tasks to learn image representations, where the objective functions are similar to those used for supervised learning. Early pretext tasks include image jigsaw puzzle~\cite{Jigsaw}, geometric transformations~\cite{Transformation}, clustering~\cite{Clustering}. Recently, contrastive learning methods, such as contrastive predictive coding~\cite{CPC}, contrastive multiview coding~\cite{CMC}, Deep InfoMax~\cite{DIM}, PIPL~\cite{PIPL}, SimCLR~\cite{SimCLR} and MOCO~\cite{MOCO}, have drawn extensive research attentions. The elfevident of these methods is to learn image representations by contrasting positive and negative pairs. 
More recently, BYOL~\cite{BYOL} takes an important step forward by using a dual online-target path paradigm and discarding negative pairs, and thus achieves superior performance over contrastive learning methods. In the dual path paradigm, an online network predicts the output of a target network and the weights of target network are updated using an exponential moving average strategy.
BYOL focuses on the image-level global consistency, while our PGL model focuses on the region-level local consistency, which enlightens the network learn complex structural representations and be beneficial to downstream medical image segmentation tasks. 

\begin{figure*}[!t]
	\begin{center}
		{\includegraphics[width=0.9\linewidth]{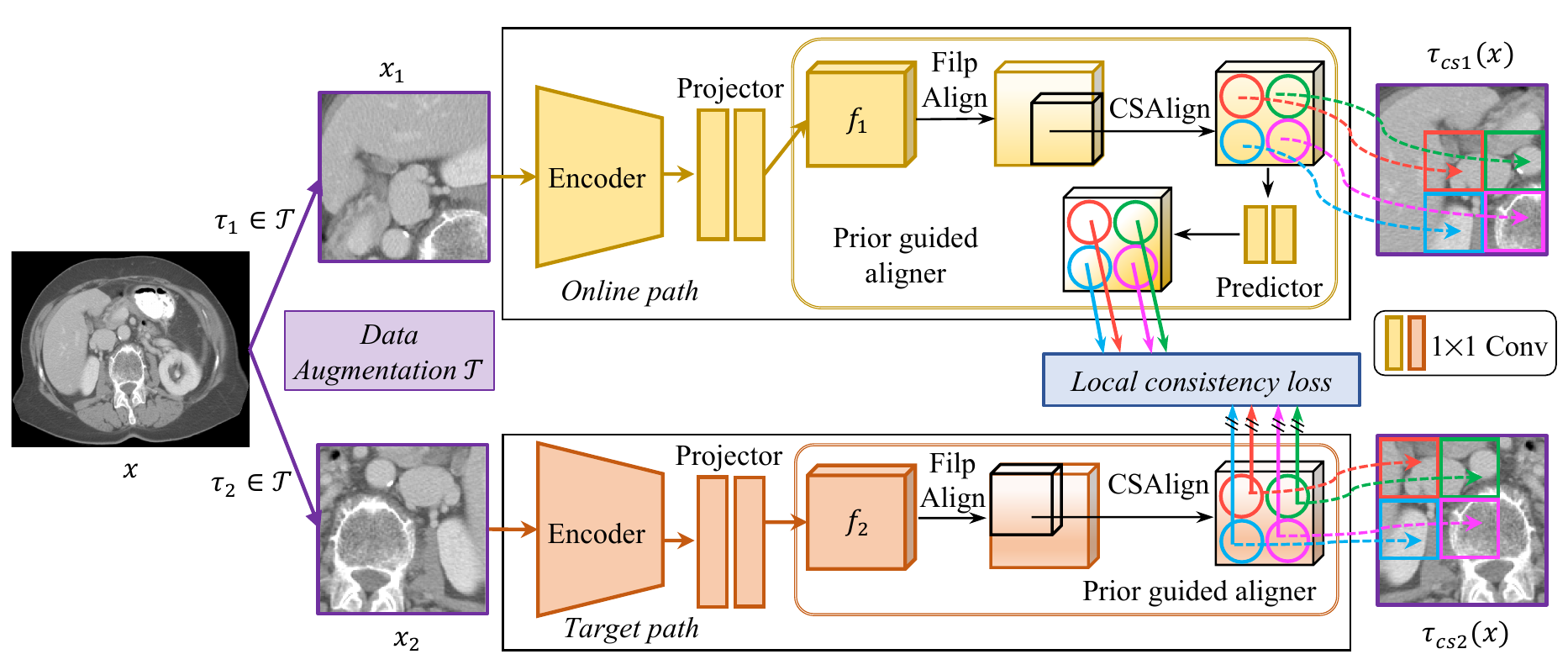}}
	\end{center}
	\caption{Illustration of proposed PGL model, including a data augmentation module $\mathcal{T}$ and a prior dual-path module. For an input image $\bm{x}$, $\mathcal{T}$ generates two views of it, denoted by $\bm{x}_1$ and $\bm{x}_2$. In the prior dual-path module, the information about augmentation transformation is introduced to the dual online-target path, aiming to align the features of the same region but extracted on different views. The PGL model uses a local consistency loss to minimize the consistency between the aligned features, and thus learn structural representations effectively. 
    The symbol \begin{scriptsize}$\backslash \backslash$\end{scriptsize} means stop-gradient.
	The $\tau_{cs}$ represents the cropping and scaling transformation. The same color rectangles within $\tau_{cs1}(\bm{x})$ and $\tau_{cs2}(\bm{x})$ cover the same region.
    }	
	\label{fig:fig2}
\end{figure*}

\section{Methods}
\subsection{Overview}
The PGL model aims to learn a universal structural representation using unlabeled volumetric medical images, and thus to improve downstream segmentation tasks.
As shown in Figure~\ref{fig:fig2}, this model consists of a data augmentation module $\mathcal{T}$ and a prior dual-path module that includes an online path and a target path.
Given an unlabeled training image $\bm{x}$, the module $\mathcal{T}$ draws two augmented views of it, denoted by $\bm{x}_1$ and $\bm{x}_2$, which are then fed to the prior dual-path module for feature extraction and alignment. 
The latent feature maps extracted by the online path and target path are denoted by $\bm{f}_1$ and $\bm{f}_2$. The transformation used in data augmentation is adopted as a prior to guide the alignment of the features, which are extracted in the same image region but located on $\bm{f}_1$ and $\bm{f}_2$, respectively.
This model uses a local consistency loss to minimize the consistency between the aligned features.
In the inference stage, we only keep the encoder in the online path for evaluation. 
We now delve into the details of this model.

\subsection{Data augmentation module}
The data augmentation module $\mathcal{T}$ transforms randomly an input image $\bm{x}$ to produce its two views
\begin{equation}
\begin{matrix}
\bm{x}_1=\tau_{1}(\bm{x}), &&
\bm{x}_2=\tau_{2}(\bm{x})
\end{matrix} 
\end{equation}
where $\tau_{1},\tau_{2} \in \mathcal{T}$. For this study, we sequentially use the following augmentation operations:

\noindent
(1) Randomly cropping and scaling. 
We randomly crop 3D patches, whose size ranges from 110\% and 140\% of the input size, from the original image.
To ensure that the transformation prior can work well in the later stage, it is required that the overlap between two views of the same input is no less than 10\%. The cropped patches are then resized to the input size.

\noindent
(2) Randomly flipping with a probability of 50\% 
along each of three axes.

\noindent
(3) Adding white Gaussian noise, whose variance is drawn from a uniform distribution between 0 and 0.1. The probability of applying this transformation is 10\%.

\noindent
(4) Gaussian blurring with a probability of 20\%.
The $\sigma$ of the Gaussian kernel is sampled from a uniform distribution between 0.5 and 1.

\noindent
(5) Adjusting rightness and contrast with a probability of 50\%. 
Voxel values are first multiplied by a constant sampled from a uniform distribution between 0.75 and 1.25, and then are clipped to the original dynamic range.

\noindent
(6) Applying the following Gamma transformation~\cite{nnUnet} on a voxel-by-voxel basis
\begin{equation}
 i_{new}=i_{old}^\lambda 
\end{equation}
where $\lambda$ is sampled from a uniform distribution between 0.7 and 1.5. The transformed voxel values are then scaled to $[0, 1]$. This transformation is applied with a probability of 50\%.

\subsection{Prior dual-path module}
The prior dual-path module contains an online path and a target path. The online path comprises an encoder $E_\theta$, a projector $P_\theta$, and a prior-guided aligner with an extra predictor $Y_\theta$, where $\bm{\theta}$ is the ensemble of weights in this path. The target path has an encoder $E_\mu$, a projector $P_\mu$, and a prior-guided aligner without a predictor. The architectures of the encoder and projector are identical to those in the online path, but the weights $\bm{\mu}$ are different.

\begin{figure}[!t]
	\begin{center}
		{\includegraphics[width=1.0\linewidth]{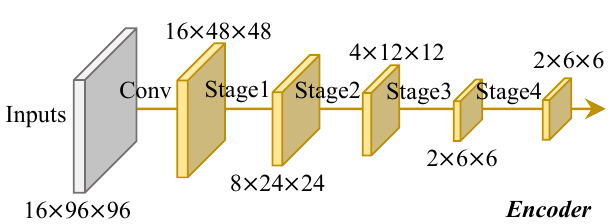}}
	\end{center}
	\vspace{-0.5cm}
	\caption{Diagram of encoder: A 3D Conv layer and four stages of 3D residual blocks. The numbers of 3D residual blocks are three, four, six, and three in four stages, respectively.
	}	
	\label{fig:fig3}
	\vspace{-0.1cm}
\end{figure}

\noindent
\textbf{Encoder.} To process volumetric data, we opt for 3D ResNet-50 as the encoder backbone, which contains a 3D convolutional (Conv) layer and four stages of 3D residual blocks, shown in Figure~\ref{fig:fig3}. The numbers of 3D residual blocks in four stages are three, four, six, and three, respectively.

\noindent
\textbf{Projector.} The projector used in~\cite{SimCLR, BYOL} maps the encoder output to the feature space where only the global consistency loss is applied, without considering the spatial information. To remedy this defect, our projector removes the global average pooling layer and replaces the two-layer perception with a Conv-BN-Relu-Conv block, which consists of a $1\times1$ Conv layer with 4096 channels followed by the batch normalization (BN), Rectified Linear Unit (ReLU) activation, and a $1\times1$ Conv layer with 256 channels.

\noindent
\textbf{Prior-guided aligner.}
Let the feature representations produced by the projectors in both pathes be formally expressed as
\begin{equation}
\begin{matrix}
\bm{f}_1=P_\theta(E_\theta(\bm{x}_1)) , &&
\bm{f}_2=P_\mu(E_\mu(\bm{x}_2))
\end{matrix}
\end{equation}
The goal of our aligner is to build the spatial relation between $\bm{f}_1$ and $\bm{f}_2$ based on the prior information of augmentation transformation and to employ such relation to align the features. 
Considering that only the random cropping / scaling and flipping can change the spatial position of an extracted patch, we propose the following two-step alignment strategy.

First, FilpAlign uses the prior of flip transformation $\tau_{F}$ to align the representations $\bm{f}_1$ and $\bm{f}_2$ to their original position as follows
\begin{equation}
\begin{matrix}
\bm{f}_{1}^{F}=\tau_{F1}^{-1}(\bm{f}_1) , &&
\bm{f}_{2}^{F}=\tau_{F2}^{-1}(\bm{f}_2)
\end{matrix} 
\end{equation}
where $\tau_{F}^{-1}$ denotes the reverse operation of $\tau_{F}$, and $\bm{f}_{1}^{F}$ and $\bm{f}_{2}^{F}$ present the outputs of online FilpAlign and target FilpAlign, respectively.

\begin{figure}[!t]
	\begin{center}
		{\includegraphics[width=1\linewidth]{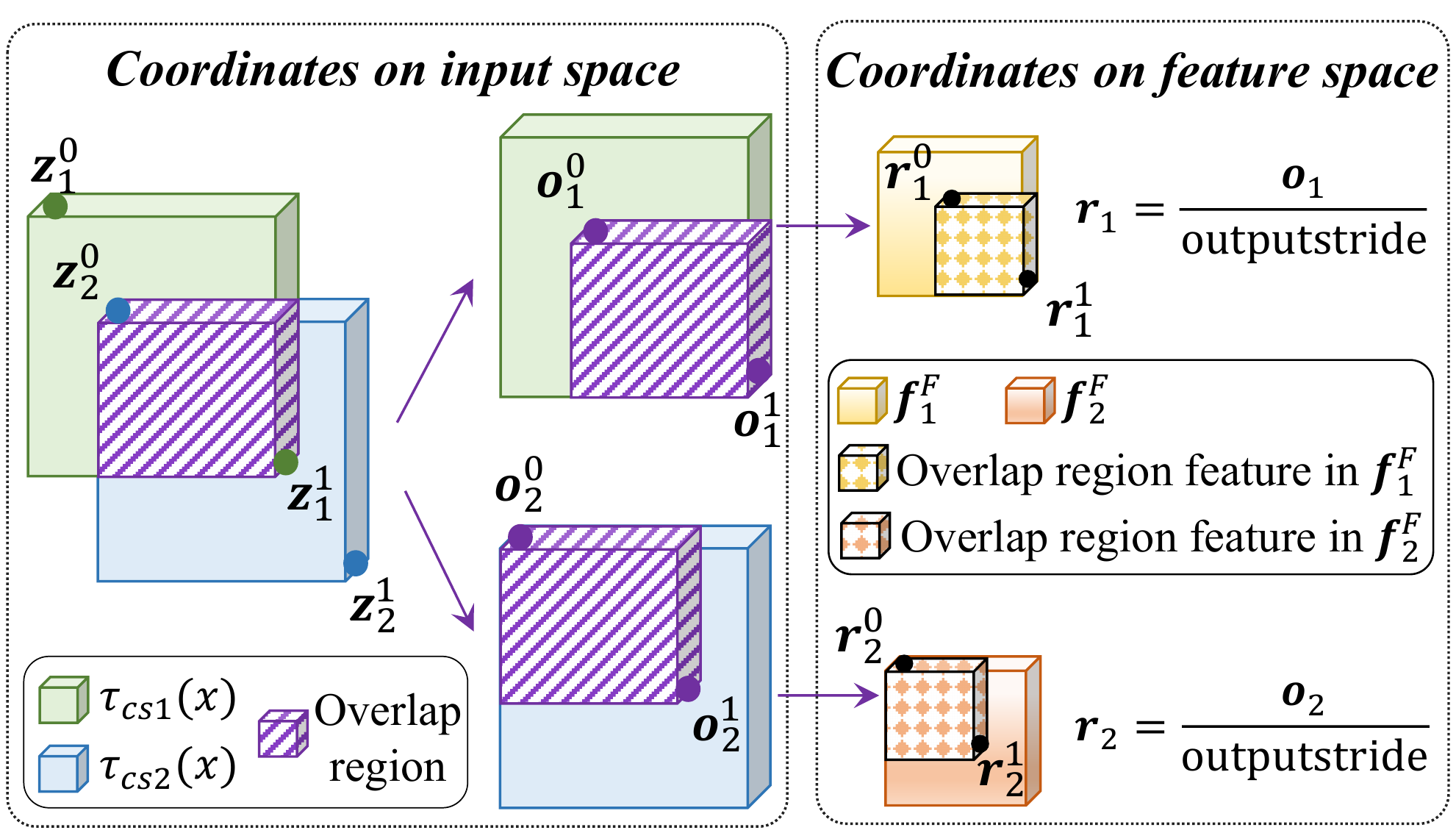}}
	\end{center}
	\vspace{-0.5cm}
	\caption{Illustration of coordinate transformation from image space to feature space, where $\bm{f}_F$ denotes the output of FilpAlign, $\tau_{cs}$ denotes the cropping and scaling transformation, and $\#^0$ and $\#^1$ denote the starting and ending coordinates, respectively.}	
	\label{fig:fig4}
    \vspace{-0.1cm}
\end{figure}

Second, CSAlign employs the prior of cropping / scaling transformation to align overlapped region in the feature space. 
Let the random cropping / scaling transformation be denoted by $\tau_{cs}$. Two views obtained by applying $\tau_{cs}$ to the input image $\bm{x}$ are denoted by $\tau_{cs1}(\bm{x})$ and $\tau_{cs2}(\bm{x})$, respectively. 
Let the starting and ending coordinates of $\tau_{cs1}(\bm{x})$ be denoted by $\bm{z}_1^0$ and $\bm{z}_1^1$, and the starting and ending coordinates of $\tau_{cs2}(\bm{x})$ be denoted by $\bm{z}_2^0$ and $\bm{z}_2^1$.
We first calculate the coordinates of the overlapped region in $\tau_{cs1}(\bm{x})$ and $\tau_{cs2}(\bm{x})$ (see Figure~\ref{fig:fig4}), shown as follows
\begin{equation}
\begin{matrix}
\bm{o}_1^0=\max(\bm{z}_1^0, \bm{z}_2^0)-\bm{z}_1^0 , & 
\bm{o}_1^1=\min(\bm{z}_1^1, \bm{z}_2^1)-\bm{z}_1^0 
\\ 
\bm{o}_2^0=\max(\bm{z}_1^0, \bm{z}_2^0)-\bm{z}_2^0 ,
& \bm{o}_2^1=\min(\bm{z}_1^1, \bm{z}_2^1)-\bm{z}_2^0 
\end{matrix} 
\end{equation}
where $\bm{o}_1^0$ and $\bm{o}_1^1$ are the starting and ending coordinates of the overlapped region in $\tau_{cs1}(\bm{x})$, and $\bm{o}_2^0$ and $\bm{o}_2^1$ are the starting and ending coordinates of the overlapped region in $\tau_{cs2}(\bm{x})$. 
Then, we perform the coordinate transformation from the image space to the feature space, shown as follows
\begin{equation}
\begin{matrix}
\bm{r}_1^0=\frac{\bm{o}_1^0}{output stride} , & \bm{r}_1^1=\frac{\bm{o}_1^1}{output stride}
\\
\bm{r}_2^0=\frac{\bm{o}_2^0}{output stride} , & \bm{r}_2^1=\frac{\bm{o}_2^1}{output stride}
\end{matrix} 
\end{equation}
where $\bm{r}_1$ and $\bm{r}_2$ are the coordinates of the overlapped region in $\bm{f}_{1}^{F}$ and $\bm{f}_{2}^{F}$, respectively, and the scaling factor $outputstride$ means the ratio between the input image size and the output feature size, which is set to ${8, 16, 16}$ in depth axis, height axis and width axis.
Next, we perform the 3D RoIAlign operation~\cite{RoIAlign} to extract the features of the overlapped region from $\bm{f}_{1}^{F}$ and $\bm{f}_{2}^{F}$ under the guidance of $\bm{r}_1$ and $\bm{r}_2$. Finally, we use the bilinear interpolation to resize extracted features to the size of $\bm{f}^{F}$. Both steps can be formally expressed as follows
\begin{equation}
\begin{matrix}
\bm{f}_{1}^{F\&CS}=R(RoIAlign(\bm{f}_1^F,\bm{r}_1)) 
\\ 
\bm{f}_{2}^{F\&CS}=R(RoIAlign(\bm{f}_2^F,\bm{r}_2))
\end{matrix} 
\end{equation}
where $R(\cdot )$ represents a resize function, and $\bm{f}_{1}^{F\&CS}$ and $\bm{f}_{2}^{F\&CS}$ are the outputs of the online CSAlign and target CSAlign, which are extracted respectively from $\bm{f}_1$ and $\bm{f}_2$ but characterizing the same region on the input image.

\subsection{Local consistency loss}
The features $\bm{f}_{1}^{F\&CS}$ and $\bm{f}_{2}^{F\&CS}$ represent the same input region, and hence ought to have a voxel-wise consistency. Accordingly, we introduce the following local consistency loss to our model. 
\begin{equation}
\mathcal{L}_{local}(\bm{x}_1, \bm{x}_2)=\frac{\sum||(\mathcal{N}(Y_\theta(\bm{f}_{1}^{F\&CS})), \mathcal{N}(\bm{f}_{2}^{F\&CS})||_2^2}{N\times H\times W\times D}
\end{equation}
where $Y_\theta$ is an additional predictor which has the same Conv-BN-Relu-Conv block as the projector, $\mathcal{N}$ means the $l_2$ normalization along the channel axis, $N$ represents the number of inputs, and $H$, $W$, and $D$ denote the height, width, and depth of the aligner features, respectively. 
Similarly, when feeding $\bm{x}_2$ to the online network and feeding $\bm{x}_1$ to the target network, the local consistency loss is defined as $\mathcal{L}_{local}(\bm{x}_2, \bm{x}_1)$. Therefore, the total loss to be minimized at each training step is shown as follows
\begin{equation}
\mathcal{L}_{total}=\mathcal{L}_{local}(\bm{x}_1, \bm{x}_2) +\mathcal{L}_{local}(\bm{x}_2, \bm{x}_1)
\end{equation}
This loss function considers the local region-level consistency between different augmented views by minimizing the voxel-wise discrepancy between the normalized online path features and target path features. Thus, our local consistency is able to assist the network in learning more structural representations than a global consistency, and hence is more suitable for downstream medical image segmentation tasks.

It is noteworthy that, since the target path just provides the ground truth signal of the online path, it does not perform gradient propagation in each training step. However, following the settings in BYOL~\cite{BYOL}, the target path weights $\bm{\mu}$ will be updated by an exponential moving average of the corresponding online path weights $\bm{\theta}$, shown as follows
\begin{equation}
\bm{\mu}=\omega * \bm{\mu}+(1-\omega )*\bm{\theta}
\end{equation}
where $\omega$ denotes the target decay rate. 

\section{Datasets}
\subsection{Pre-text task dataset.}
To train our PGL model in a self-supervised manner, we collected 1808 CT scans from 5 public datasets, including 660 CT scans from the RibFrac dataset~\footnote{\url{https://ribfrac.grand-challenge.org/}}~\cite{Ribfrac_data} and 1148 CT scans from four subsets (\textit{i.e.}, Hepatic Vessel, Colon Tumor, Pancreas, and Lung Tumor) of the Medical Segmentation Decathlon (MSD) challenge~\footnote{\url{http://medicaldecathlon.com/}}~\cite{MSD_data}.

\subsection{Downstream task datasets.}
Four public datasets were used to evaluate our PGL model on downsteam tasks.

\noindent
\textbf{The Liver dataset} is from the MSD challenge. It contains 131 labeled CT scans acquired at seven hospitals and research centers, each being paired with the liver and liver tumour boundaries traced manually by radiologists. This dataset was split randomly into a training set of 66 scans and a test set of 65 scans.

\noindent
\textbf{The Spleen dataset} is from the MSD challenge. It includes 41 labeled CT scans, in which the spline was drawn using a level-set based approach and the contours were manually adjusted by expert abdominal radiologists. This dataset was split randomly into a training set of 20 scans and a test set of 21 scans.

\noindent
\textbf{The KiTS dataset~\footnote{\url{https://kits19.grand-challenge.org/home/}}}~\cite{KiTS_data} consists of 210 labeled CT scans for automatic kidney and kidney tumour segmentation. Each scan is annotated by medical students under the supervision of their clinical chair. This dataset was split equally and randomly into a training set and a test set.

\noindent
\textbf{The BCV dataset~\footnote{\url{https://www.synapse.org/\#!Synapse:syn3193805/wiki/217789}}} comprises 30 labeled CT scans for automatic segmentation of 13 abdominal organs, including the spleen (Sp), right kidney (RK), left kidney (LK), gallbladder (Gb), esophagus (Es), liver (Li), stomach (St), aorta, inferior vena cava (IVC), portal vein and splenic vein (PSV), pancreas (Pa), right adrenal gland (RAG), and left adrenal gland (LAG). This dataset was split equally and randomly into a training set and a test set.

\section{Experiments and Results}
\subsection{Implementation details}
\noindent
\textbf{Training PGL model.}
For each CT scan, we first clipped voxel values to the range [-1024, +325]HU that covers most organs and tumors, and then normalized truncated voxel values by subtracting their mean and dividing by their standard deviation.
We set the size of input patches to $16\times96\times96$, aiming to weigh the balance between reserving enough information for self-supervision and reducing computational and spatial complexity to an affordable level.
Following~\cite{BYOL}, we adopted the LARS optimizer~\cite{LARS} with a cosine decaying learning rate~\cite{cosine_LR}, a warm-up period of 2 epochs, and no restart. We set the initial learning rate to 0.2, batch size to 128, and maximum epochs to 500. For the target path, the decay rate $\omega$ increases from 0.996 to 1.000 during training~\cite{BYOL}.

\noindent
\textbf{Segmentation network for downstream task.}
The segmentation network for downstream tasks has a U-like architecture~\cite{Unet}, consisting of a 3D ResNet-50 encoder, an atrous spatial pyramid pooling (ASPP) component~\cite{ASPP}, a decoder and three skip connections between the encoder and decoder. ASPP has three 3D separable Conv (SepConv) layers with dilation rates of 2, 4, and 8, respectively. The decoder contains four transposed convolutional (TransConv) blocks, and a Conv layer for prediction. Each TransConv block consists of a TransConv layer and a residual block with two SepConv layers.
Note that the segmentation network is designed with a decoder-light architecture. The parameters of ASPP and decoder only account for 5\% of the total parameters (2.38 millions \textit{vs.} 48.54 millions). The 3D ResNet-50 encoder was pre-trained by our PGL model, and other parts of the segmentation network are randomly initialized. We jointly employed the Dice loss and cross-entropy loss as the object function for optimization, which is popular in many medical image segmentation applications and achieves prominent success~\cite{Loss_Zhang, Unet++}.
Additional details on the segmentation network are in Appendix. 

\noindent
\textbf{Fine-tuning segmentation network.}
We employed the data augmentation techniques used for training the PGL model to enlarge the downstream training dataset.
The input size was set to $48\times160\times160$ on the BCV dataset and $64\times192\times192$ on other datasets.
The stochastic gradient descent (SGD) algorithm with a batch size of 8 was adopted to optimize the segmentation network. We set the initialial learning rate to 0.01 and the maximum epochs to 100. We randomly selected 20\% of training scans to form a validation set, which was used to monitor the performance of segmentation network. We will terminate the training process when the network falls into overfitting.

\noindent
\textbf{Performance metrics.}
For this study, the segmentation performance of each method is measured by the Dice coefficient scores, Intersection over Union (IoU), and Hausdorff distance (HD).

\subsection{Comparison to state of the art}
We compared the proposed PGL pre-training strategy with two state-of-the-art pre-training strategies, \textit{i.e.}, the Models Genesis~\cite{ZZW} and BYOL~\cite{BYOL}, on four datasets. The baseline segmentation network was randomly initialized using the Kaiming uniform method~\cite{KaimingIni}. 
The Models Genesis~\cite{ZZW} is the most famous SSL method for medical image analysis. In this method, sub-volumes are first cropped from CT scans and then deformed using up to three out of four transformations, including non-linear, local-shuffling, outer-cutout and inner-cutout. Next, a deep learning model is trained to restore the deformed sub-volumes to their primeval state. The learned weights are used to initialize the downstream segmentation network.
As aforementioned, BYOL learns image representations using a dual online-target path paradigm with a global consistency constraint and is one of the most successful SSL methods for many vision applications.
For a fair comparison, BYOL and our PGL model used the same encoder (\textit{i.e.}, 3D ResNet-50) and inputs in the self-supervised training stage. The Models Genesis used the same architecture to our segmentation network.
In the fine-tuning stage, all methods used the same segmentation network and inputs.

The performance of the baseline segmentation network using either random initialization or one of those three pre-training strategies on four datasets was displayed in Table~\ref{tab:tab1} to Table~\ref{tab:tab4}. 
It shows that initialization with our PGL model elevates consistently the segmentation performance over the random initialization on four datasets, improving the average Dice by 2.08\% on the Liver dataset, 2.37\% on the Spleen dataset, 2.72\% on KiTS, and 2.2\% on BCV. Particularly for the multi-organ segmentation task on BCV, using PGL to initialize the segmentation network substantially improves Dice scores of gallbladder and stomach segmentation by 7.93\% and 4.25\%, respectively. The results suggest that our PGL model has the potential to provide a generalizable and transferable initialization to improve the performance of 3D medical image segmentation with limited annotations.
Moreover, comparing to using the Models Genesis or BYOL, using our PGL model for initialization enables the segmentation network to be more accurate. Specifically, using PGL improves the average Dice by 1.23\% on the Liver dataset, 1.04\% on the Spleen dataset, 0.23\% on KiTS, and 0.74\% on BCV over the second best pre-training strategy (\textit{i.e.}, BYOL).
The superior performance of our PGL model corroborates that our local consistency learning strategy has a stronger ability than the global consistency strategy used in BYOL to learn effective representations from unlabeled data for medical image segmentation.

\begin{table*}[]
\small
\caption{Performance of baseline segmentation network using either random initialization (Random Init) or one of three pre-training strategies (\textit{i.e.}, Models Genesis (MG), BYOL, and our PGL model) on BCV dataset. 'Ave': Average performance in segmentation of 13 organs.}
\label{tab:tab1}
\vspace{-0.6cm}
\begin{center}
\begin{tabular}{m{0.8cm}<{\centering}|m{1.6cm}<{\centering}|m{0.58cm}<{\centering}|m{0.58cm}<{\centering}|m{0.58cm}<{\centering}|m{0.58cm}<{\centering}|m{0.58cm}<{\centering}|m{0.58cm}<{\centering}|m{0.58cm}<{\centering}|m{0.58cm}<{\centering}|m{0.58cm}<{\centering}|m{0.58cm}<{\centering}|m{0.58cm}<{\centering}|m{0.58cm}<{\centering}|m{0.58cm}<{\centering}|m{0.58cm}<{\centering}}
\hline
\multicolumn{2}{c|}{}                                 & \multicolumn{13}{c|}{Organs}                                                                           &                       \\ \cline{3-15}
\multicolumn{2}{c|}{\multirow{-2}{*}{Methods}}        & Sp    & RK    & LK    & Gb    & Es    & Li    & St    & Aorta & IVC   & PSV & Pa    & RAG   & LAG   & \multirow{-2}{*}{Ave} \\ \hline
\rowcolor[HTML]{F8EAE9} 
\cellcolor[HTML]{F8EAE9}                       & Random Init & 94.01 & 92.97 & 92.15 & 51.98 & 71.85 & 94.82 & 77.74 & 87.47 & 84.85 & 70.91  & 74.12 & 62.27 & 67.30  & 78.65                 \\ \cline{2-16} 
\rowcolor[HTML]{F8EAE9} 
\cellcolor[HTML]{F8EAE9}                       & MG~\cite{ZZW}   & 94.92 & 93.03 & 91.87 & 59.80 & 71.28 & 95.27 & 80.88 & 87.92 & 85.34 & 71.95  & 75.88 & 63.70 & 67.77 & 79.97                 \\ \cline{2-16} 
\rowcolor[HTML]{F8EAE9} 
\cellcolor[HTML]{F8EAE9}                       & BYOL~\cite{BYOL} & 95.04 & 93.53 & 92.55 & 59.70 & 70.98 & 95.35 & 80.69 & 88.37 & 85.36 & 71.93  & 75.95 & 63.71 & 68.27 & 80.11                 \\ \cline{2-16} 
\rowcolor[HTML]{F8EAE9} 
\multirow{-4}{*}{\cellcolor[HTML]{F8EAE9}Dice~$\uparrow$} & \textbf{PGL(Ours)} & \textbf{95.46} & \textbf{93.54} & \textbf{92.62} & \textbf{59.91} & \textbf{72.59} & \textbf{96.14} & \textbf{81.99} & \textbf{89.20} & \textbf{86.49} & \textbf{72.50}  & \textbf{77.00} & \textbf{63.85} & \textbf{69.75} & \textbf{80.85}                 \\ \hline
\rowcolor[HTML]{EBECFE} 
\cellcolor[HTML]{EBECFE}                       & Random Init & 88.87 & 86.95 & 85.76 & 40.97 & 57.12 & 90.51 & 65.87 & 78.65 & 73.95 & 55.42  & 59.48 & 46.95 & 51.43 & 67.84                 \\ \cline{2-16} 
\rowcolor[HTML]{EBECFE} 
\cellcolor[HTML]{EBECFE}                       & MG~\cite{ZZW}   & 90.44 & 87.06 & 85.19 & \textbf{47.86} & 56.52 & 91.28 & 69.98 & 79.07 & 74.68 & 56.74  & 61.60 & 48.31 & 51.8  & 69.27                 \\ \cline{2-16} 
\rowcolor[HTML]{EBECFE} 
\cellcolor[HTML]{EBECFE}                       & BYOL~\cite{BYOL} & 90.63 & 87.89 & 86.42 & 47.73 & 56.30 & 91.43 & 69.71 & 79.74 & 74.72 & 56.70  & 61.69 & 48.06 & 52.31 & 69.49                 \\ \cline{2-16} 
\rowcolor[HTML]{EBECFE} 
\multirow{-4}{*}{\cellcolor[HTML]{EBECFE}IoU~$\uparrow$}  & \textbf{PGL(Ours)} & \textbf{91.35} & \textbf{87.93} & \textbf{86.50} & 47.72 & \textbf{58.19} & \textbf{92.63} & \textbf{71.84} & \textbf{80.90} & \textbf{76.38} & \textbf{57.37}  & \textbf{63.00} & \textbf{48.32} & \textbf{54.16} & \textbf{70.48}                 \\ \hline
\rowcolor[HTML]{FEFED6} 
\cellcolor[HTML]{FEFED6}                       & Random Init & 38.31 & 2.06  & 2.54  & 51.75 & 8.83  & 3.64  & 48.28 & 26.92 & 6.12  & 16.73  & 14.66 & 5.22  & 3.82  & 17.61                 \\ \cline{2-16} 
\rowcolor[HTML]{FEFED6} 
\cellcolor[HTML]{FEFED6}                       & MG~\cite{ZZW}   & 4.43  & 2.07  & 24.89 & 12.69 & 7.65  & 3.24  & 20.77 & 26.20 & 5.20  & \textbf{8.61}   & 6.02  & 5.31  & 4.44  & 10.12                 \\ \cline{2-16} 
\rowcolor[HTML]{FEFED6} 
\cellcolor[HTML]{FEFED6}                       & BYOL~\cite{BYOL} & 3.46  & 1.92  & \textbf{2.45}  & 25.96 & 20.41 & 3.11  & 22.61 & 17.45 & 5.20  & 16.36  & 5.94  & \textbf{4.52}  & 4.46  & 10.30                 \\ \cline{2-16} 
\rowcolor[HTML]{FEFED6} 
\multirow{-4}{*}{\cellcolor[HTML]{FEFED6}HD~$\downarrow$} & \textbf{PGL(Ours)} & \textbf{2.50}  & \textbf{1.83}  & 2.47  & \textbf{11.52} & \textbf{7.18}  & \textbf{2.52}  & \textbf{12.45} & \textbf{6.23}  & \textbf{4.77}  & 13.85  & \textbf{6.00}  & 4.75  & \textbf{3.56}  & \textbf{6.13}                  \\ \hline
\end{tabular}
\end{center}
\vspace{-0.3cm}
\end{table*}

\begin{table}[]
\small
\caption{Performance of baseline segmentation network using either random initialization (Random Init) or one of three pre-training strategies on Liver dataset. 'Ave': Average performance in liver and liver tumor segmentation.}
\label{tab:tab2}
\vspace{-0.6cm}
\begin{center}
\begin{tabular}{m{0.8cm}<{\centering}|m{0.8cm}<{\centering}|m{1cm}<{\centering}|m{1.6cm}<{\centering}|m{0.8cm}<{\centering}|m{0.8cm}<{\centering}}
\hline
\multicolumn{2}{c|}{}         & \begin{tabular}[c]{@{}c@{}}Random \\ {Init}\end{tabular}  & \begin{tabular}[c]{@{}c@{}}Models \\ {Genesis~\cite{ZZW}}\end{tabular}    & \begin{tabular}[c]{@{}c@{}}BYOL \\ {~\cite{BYOL}}\end{tabular}  & \begin{tabular}[c]{@{}c@{}}\textbf{PGL} \\ {\textbf{(Ours)}}\end{tabular}  \\ \hline
\multirow{3}{*}{Organ} & Dice~$\uparrow$ & 95.73  & 96.00 & 96.29 & \textbf{96.43} \\ \cline{2-6} 
                       & IoU~$\uparrow$  & 91.94  & 92.45 & 92.92 & \textbf{93.16} \\ \cline{2-6} 
                       & HD~$\downarrow$ & 7.49   & 4.75  & 5.46  & \textbf{4.72}  \\ \hline
\multirow{3}{*}{Tumor} & Dice~$\uparrow$ & 52.20  & 53.47 & 53.34 & \textbf{55.66} \\ \cline{2-6} 
                       & IoU~$\uparrow$  & 41.64  & 42.91 & 43.25 & \textbf{44.95} \\ \cline{2-6} 
                       & HD~$\downarrow$ & 29.69  & 29.87 & 32.82 & \textbf{24.01} \\ \hline
\multirow{3}{*}{Ave}   & Dice~$\uparrow$ & 73.97 & 74.74 & 74.82 & \textbf{76.05} \\ \cline{2-6} 
                       & IoU~$\uparrow$  & 66.79  & 67.68 & 68.09 & \textbf{69.06} \\ \cline{2-6} 
                       & HD~$\downarrow$ & 18.59  & 17.31 & 19.14 & \textbf{14.37} \\ \hline

\end{tabular}
\end{center}
\vspace{-0.2cm}
\end{table}

\begin{table}[]
\small
\caption{Performance of baseline segmentation network using either random initialization (Random Init) or one of three pre-training strategies on KiTS dataset. 'Ave': Average performance in kidney and kidney tumor segmentation.}
\label{tab:tab3}
\vspace{-0.6cm}
\begin{center}
\begin{tabular}{m{0.8cm}<{\centering}|m{0.8cm}<{\centering}|m{1cm}<{\centering}|m{1.6cm}<{\centering}|m{0.8cm}<{\centering}|m{0.8cm}<{\centering}}
\hline
\multicolumn{2}{c|}{}         & \begin{tabular}[c]{@{}c@{}}Random \\ {Init}\end{tabular}  & \begin{tabular}[c]{@{}c@{}}Models \\ {Genesis~\cite{ZZW}}\end{tabular}    & \begin{tabular}[c]{@{}c@{}}BYOL \\ {~\cite{BYOL}}\end{tabular}  & \begin{tabular}[c]{@{}c@{}}\textbf{PGL} \\ {\textbf{(Ours)}}\end{tabular}  \\ \hline
\multirow{3}{*}{Organ} & Dice$\uparrow$ & 96.07 & 96.29    & 96.23 & \textbf{96.80}     \\ \cline{2-6} 
                       & IoU~$\uparrow$  & 92.62 & 92.94    & 92.99 & \textbf{93.83}     \\ \cline{2-6} 
                       & HD~$\downarrow$ & 1.95  & 1.84     & 2.1   & \textbf{1.34}      \\ \hline
\multirow{3}{*}{Tumor} & Dice~$\uparrow$ & 67.07 & 68.35    & \textbf{71.89} & 71.77     \\ \cline{2-6} 
                       & IoU~$\uparrow$  & 56.63 & 58.07    & 62.12 & \textbf{62.70}     \\ \cline{2-6} 
                       & HD~$\downarrow$ & 25.64 & 21.12    & 21.73 & \textbf{17.59}     \\ \hline
\multirow{3}{*}{Ave}   & Dice~$\uparrow$ & 81.57 & 82.32    & 84.06 & \textbf{84.29}     \\ \cline{2-6} 
                       & IoU~$\uparrow$  & 74.63 & 75.51    & 77.56 & \textbf{78.27}     \\ \cline{2-6} 
                       & HD~$\downarrow$ & 13.80 & 11.48    & 11.92 & \textbf{9.47}      \\ \hline

\end{tabular}
\end{center}
\vspace{-0.4cm}
\end{table}

\begin{table}[]
\small
\caption{Performance of baseline segmentation network using either random initialization (Random Init) or one of three pre-training strategies on Spleen dataset.}
\label{tab:tab4}
\vspace{-0.5cm}
\begin{center}
\begin{tabular}{cccc}
\hline
\multicolumn{1}{c|}{Methods}   & \multicolumn{1}{c|}{Dice~$\uparrow$}  & \multicolumn{1}{c|}{IoU~$\uparrow$}   & HD~$\downarrow$ \\ \hline
\multicolumn{1}{c|}{Random init}      & \multicolumn{1}{c|}{93.23} & \multicolumn{1}{c|}{88.08} & 4.30 \\ \hline
\multicolumn{1}{c|}{Models Genesis~\cite{ZZW}}       & \multicolumn{1}{c|}{94.20}      & \multicolumn{1}{c|}{89.44}      &  5.89    \\ \hline
\multicolumn{1}{c|}{BYOL~\cite{BYOL}}      & \multicolumn{1}{c|}{94.56} & \multicolumn{1}{c|}{89.83} & 3.08 \\ \hline
\multicolumn{1}{c|}{\textbf{PGL(Ours)}} & \multicolumn{1}{c|}{\textbf{95.60}} & \multicolumn{1}{c|}{\textbf{91.61}} & \textbf{2.14} \\ \hline
\end{tabular}
\end{center}
\vspace{-0.1cm}
\end{table}

\begin{table*}[]
\small
\caption{Performance of baseline segmentation network initialized using our PGL model with different spatial transformations on BCV dataset. 'Ave': Average performance in segmentation of 13 organs.}
\label{tab:tab5}
\vspace{-0.6cm}
\begin{center}
\begin{tabular}{m{0.8cm}<{\centering}|m{1.6cm}<{\centering}|m{0.58cm}<{\centering}|m{0.58cm}<{\centering}|m{0.58cm}<{\centering}|m{0.58cm}<{\centering}|m{0.58cm}<{\centering}|m{0.58cm}<{\centering}|m{0.58cm}<{\centering}|m{0.58cm}<{\centering}|m{0.58cm}<{\centering}|m{0.58cm}<{\centering}|m{0.58cm}<{\centering}|m{0.58cm}<{\centering}|m{0.58cm}<{\centering}|m{0.58cm}<{\centering}}
\hline
\multicolumn{2}{c|}{}                                       & \multicolumn{13}{c|}{Organs}                                                                          &                       \\ \cline{3-15}
\multicolumn{2}{c|}{\multirow{-2}{*}{Methods}}              & Sp    & RK    & LK    & Gb    & Es    & Li    & St    & Aorta & IVC   & PSV   & Pa    & RAG   & LAG   & \multirow{-2}{*}{Ave} \\ \hline
\rowcolor[HTML]{F8EAE9} 
\cellcolor[HTML]{F8EAE9}                       & w/o F\&CS  & 94.87 & 92.80 & 91.57 & 55.06 & 70.90 & 94.03 & 79.41 & 86.76 & 85.51 & 70.88 & 74.55 & 61.38 & 65.43 & 78.70                 \\ \cline{2-16} 
\rowcolor[HTML]{F8EAE9} 
\cellcolor[HTML]{F8EAE9}                       & w/o CS     & 95.21 & 93.15 & 91.79 & 55.66 & 71.99 & 95.32 & 79.58 & 86.94 & 85.34 & 72.07 & 75.36 & 63.08 & 66.78 & 79.41                 \\ \cline{2-16} 
\rowcolor[HTML]{F8EAE9} 
\cellcolor[HTML]{F8EAE9}                       & w/o F      & 95.05 & 93.43 & 92.26 & \textbf{61.69} & 71.97 & 95.59 & 79.60 & 87.43 & 86.09 & 72.26 & 75.53 & \textbf{65.29} & 68.66 & 80.37                 \\ \cline{2-16} 
\rowcolor[HTML]{F8EAE9} 
\multirow{-4}{*}{\cellcolor[HTML]{F8EAE9}Dice~$\uparrow$} & \textbf{with F\&CS} & \textbf{95.46} & \textbf{93.54} & \textbf{92.62} & 59.91 & \textbf{72.59} & \textbf{96.14} & \textbf{81.99} & \textbf{89.20} & \textbf{86.49} & \textbf{72.50} & \textbf{77.00} & 63.85 & \textbf{69.75} & \textbf{80.85}                 \\ \hline
\rowcolor[HTML]{EBECFE} 
\cellcolor[HTML]{EBECFE}                       & w/o F\&CS  & 90.36 & 86.67 & 85.11 & 42.97 & 56.05 & 89.72 & 68.93 & 77.62 & 74.93 & 55.62 & 59.98 & 46.21 & 49.85 & 68.00                 \\ \cline{2-16} 
\rowcolor[HTML]{EBECFE} 
\cellcolor[HTML]{EBECFE}                       & w/o CS     & 90.89 & 87.24 & 85.32 & 44.56 & 57.23 & 91.31 & 68.74 & 77.86 & 74.60 & 56.91 & 61.14 & 47.67 & 50.95 & 68.80                 \\ \cline{2-16} 
\rowcolor[HTML]{EBECFE} 
\cellcolor[HTML]{EBECFE}                       & w/o F      & 90.65 & 87.73 & 86.01 & \textbf{49.30} & 57.12 & 91.73 & 68.80 & 78.63 & 75.81 & 57.25 & 61.27 & \textbf{49.81} & 53.02 & 69.78                 \\ \cline{2-16} 
\rowcolor[HTML]{EBECFE} 
\multirow{-4}{*}{\cellcolor[HTML]{EBECFE}IoU~$\uparrow$}  & \textbf{with F\&CS} & \textbf{91.35} & \textbf{87.93} & \textbf{86.50}  & 47.72 & \textbf{58.19} & \textbf{92.63} & \textbf{71.84} & \textbf{80.90} & \textbf{76.38} & \textbf{57.37} & \textbf{63.00} & 48.32 & \textbf{54.16} & \textbf{70.48}                 \\ \hline
\rowcolor[HTML]{FEFED6} 
\cellcolor[HTML]{FEFED6}                       & w/o F\&CS  & 3.53  & 2.01  & 2.90  & 16.84 & 8.88  & 3.69  & 17.57 & 27.14 & 5.31  & 22.54 & 8.89  & \textbf{3.49}  & 5.02  & 9.83                  \\ \cline{2-16} 
\rowcolor[HTML]{FEFED6} 
\cellcolor[HTML]{FEFED6}                       & w/o CS     & \textbf{2.46}  & 2.01  & 3.10  & 18.24 & 9.12  & 2.88  & 19.99 & 27.10 & 4.97  & 15.49 & 6.56  & 5.63  & 4.40  & 9.38                  \\ \cline{2-16} 
\rowcolor[HTML]{FEFED6} 
\cellcolor[HTML]{FEFED6}                       & w/o F      & 3.58  & 1.86  & 2.53  & \textbf{10.37} & 7.96  & 2.87  & 21.21 & 26.98 & 5.06  & 18.45 & 13.97 & 4.79  & 4.18  & 9.52                  \\ \cline{2-16} 
\rowcolor[HTML]{FEFED6} 
\multirow{-4}{*}{\cellcolor[HTML]{FEFED6}HD~$\downarrow$} & \textbf{with F\&CS} & 2.50  & \textbf{1.83}  & \textbf{2.47}  & 11.52 & \textbf{7.18}  & \textbf{2.52}  & \textbf{12.45} & \textbf{6.23}  & \textbf{4.77}  & \textbf{13.85} & \textbf{6.00}  & 4.75  & \textbf{3.56}  & \textbf{6.16}                  \\ \hline
\end{tabular}

\end{center}
\vspace{-0.3cm}
\end{table*}

\begin{table}[]
\small
\caption{Segmentation performance of our PGL model with different spatial transformations on Liver dataset. 'Ave': Average performance in liver and liver tumor segmentation.}
\label{tab:tab6}
\vspace{-0.6cm}
\begin{center}
\begin{tabular}{m{0.9cm}<{\centering}|m{0.9cm}<{\centering}|m{1.cm}<{\centering}|m{1.1cm}<{\centering}|m{0.9cm}<{\centering}|m{1cm}<{\centering}}
\hline
\multicolumn{2}{c|}{}         & w/o F\&CS & w/o CS & w/o F & \textbf{with F\&CS} \\ \hline
\multirow{3}{*}{Organ} & Dice~$\uparrow$ & 96.06     & 96.12  & 96.40  &\textbf{96.43}      \\ \cline{2-6} 
                       & IoU~$\uparrow$  & 92.47     & 92.61  & 93.10 & \textbf{93.16}      \\ \cline{2-6} 
                       & HD~$\downarrow$ & 6.13      & 5.53   & 4.89  & \textbf{4.72}       \\ \hline
\multirow{3}{*}{Tumor} & Dice~$\uparrow$ & 52.50     & 54.10  & 55.30 & \textbf{55.66}      \\ \cline{2-6} 
                       & IoU~$\uparrow$  & 42.03     & 43.58  & 44.69 & \textbf{44.95}      \\ \cline{2-6} 
                       & HD~$\downarrow$ & 32.83     & 27.09  & 29.51 & \textbf{24.01}      \\ \hline
\multirow{3}{*}{Ave}   & Dice~$\uparrow$ & 74.28     & 75.11  & 75.85 & \textbf{76.05}      \\ \cline{2-6} 
                       & IoU~$\uparrow$  & 67.25     & 68.10  & 68.90 & \textbf{69.06}      \\ \cline{2-6} 
                       & HD~$\downarrow$ & 19.48     & 16.31  & 17.2  & \textbf{14.37}      \\ \hline
\end{tabular}
\end{center}
\vspace{-0.2cm}
\end{table}

\begin{table}[]
\small
\caption{Segmentation performance of our PGL model with different spatial transformations on KiTS dataset. 'Ave': Average performance in kidney and kidney tumor segmentation.}
\label{tab:tab7}
\vspace{-0.6cm}
\begin{center}
\begin{tabular}{m{0.9cm}<{\centering}|m{0.9cm}<{\centering}|m{1.cm}<{\centering}|m{1.1cm}<{\centering}|m{0.9cm}<{\centering}|m{1cm}<{\centering}}
\hline
\multicolumn{2}{c|}{}        & w/o F\&CS & w/o CS & w/o F & \textbf{with F\&CS} \\ \hline
\multirow{3}{*}{Organ} & Dice~$\uparrow$ & 96.17     & 96.34  & 96.54 & \textbf{96.80}      \\ \cline{2-6} 
                       & IoU~$\uparrow$  & 92.72     & 93.01  & 93.36 & \textbf{93.83}      \\ \cline{2-6} 
                       & HD~$\downarrow$ & 3.47      & 1.75   & 1.59  & \textbf{1.34}       \\ \hline
\multirow{3}{*}{Tumor} & Dice~$\uparrow$ & 67.72     & 68.34  & 68.76 & \textbf{71.77}      \\ \cline{2-6} 
                       & IoU~$\uparrow$  & 57.62     & 57.86  & 59.28 & \textbf{62.70}      \\ \cline{2-6} 
                       & HD~$\downarrow$ & 19.89     & 22.04  & \textbf{15.64} & 17.59      \\ \hline
\multirow{3}{*}{Ave}   & Dice~$\uparrow$ & 81.95     & 82.34  & 82.65 & \textbf{84.29}      \\ \cline{2-6} 
                       & IoU~$\uparrow$  & 75.17     & 75.44  & 76.32 & \textbf{78.27}      \\ \cline{2-6} 
                       & HD~$\downarrow$ & 11.68     & 11.90  & \textbf{8.62}  & 9.47       \\ \hline

\end{tabular}
\end{center}
\vspace{-0.5cm}
\end{table}

\begin{table}[]
\small
\caption{Segmentation performance of our PGL model with different spatial transformations on Spleen dataset.}
\label{tab:tab8}
\vspace{-0.5cm}
\begin{center}
\begin{tabular}{c|c|c|c}
\hline
Methods    & Dice~$\uparrow$  & IoU~$\uparrow$   & HD~$\downarrow$ \\ \hline
w/o F\&CS  & 92.41 & 87.03 & 5.43 \\ \hline
w/o CS     & 94.16 & 89.32 & 2.98 \\ \hline
w/o F      & 93.53 & 88.44 & 5.48 \\ \hline
\textbf{with F\&CS (Ours)} & \textbf{95.60} & \textbf{91.61} & \textbf{2.14} \\ \hline
\end{tabular}
\end{center}
\vspace{-0.5cm}
\end{table}

\subsection{Discussions}
\noindent
\textbf{Effectiveness of prior spatial transformations.}
In PGL, we employ the prior of spatial transformations (\textit{i.e.}, flipping and cropping / scaling) to boost the model's ability to learn structural representations from unlabeled data. 
To evaluate the effectiveness of this operation, we compared the segmentation performance of using four settings of our PGL model: (1) without using both transformations (denoted by 'w/o F\&CS') and thus removing both FilpAlign and CSAlign, (2) without using cropping / scaling (denoted by 'w/o CS') and removing CSAlign, (3) without using flipping (denoted by 'w/o F') and removing FilpAlign, (4) the proposed method that uses both transformations (denoted by 'with F\&CS').
The segmentation performance obtained on four datasets was shown in Table~\ref{tab:tab5} to Table~\ref{tab:tab8}. 
It reveals that (1) if both transformations and the prior information were not used, it is difficult for PGL to learn complex structural representations from unlabeled data, leading to the lowest performance, even worse than random initialization; (2) the cropping / scaling transformation in our PGL is more important than flipping; and (3) jointly using cropping / scaling and flipping enables the model to produce more accurate segmentation than using either transformation alone.

\begin{figure}[!t]
	\begin{center}
		{\includegraphics[width=1\linewidth]{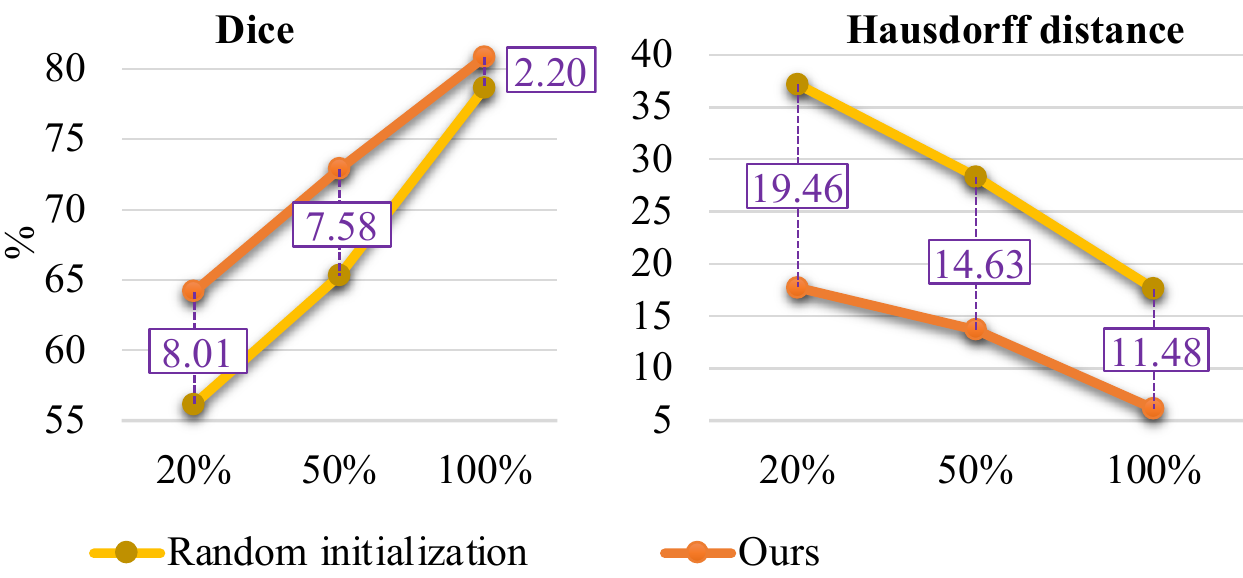}}
	\end{center}
	\vspace{-0.5cm}
	\caption{Performance of baseline segmentation network using random initialization or our PGL-based pre-training on BCV dataset, when using 20\%, 50\%, or 100\% labeled downstream training images.}	
	\label{fig:fig5}
	\vspace{-0.3cm}
\end{figure}
 
\noindent
\textbf{Robustness to less annotations.}
Our PGL model can learn from unlabeled images and transfer the learned representation ability to the downstream segmentation tasks. Therefore using the pre-trained PGL model to initialize a downstream network leads to more robust performance than training the network from scratch, particularly when the annotated training dataset is small. 
To validate this, we kept the test set and pre-trained weights unchanged, and randomly selected 20\% and 50\% scans from the BCV training set, respectively, to fine-tune the segmentation network. 
As a control, we also used those selected data to train the randomly initialized segmentation network. 
The average segmentation performance on 13 organs obtained in this experiment was depicted in Figure~\ref{fig:fig5}. It shows that the initialization with pre-trained PGL model beats steadily random initialization, no matter how many labeled scans were used for fine-tuning. Moreover, it suggests that our PGL model can alleviate, to some extent, the lack of labeled training data, leading to larger performance gain over random initialization when the labeled training dataset is smaller. 
Additional details on the segmentation results of 13 organs are in Appendix.

\noindent
\textbf{Computational complexity.}
The PGL model was trained using a workstation with four NVIDIA GTX 2080Ti GPUs and 128GB Memory, and downstream tasks were conducted on a workstation with two 2080Ti GPUs, both using the Pytorch software packages and distributed training. It took about 5 days to train the PGL model, about 2 days to train the segmentation network for each downstream task, and about 30ms to segment a volume of size $64\times192\times192$.

\section{Conclusion}
This paper proposes the PGL self-supervised model to learn universal representations from unlabeled 3D medical images for downstream 3D medical image segmentation. In this model, the spatial transformation that different views of the same image is used as a prior to guide the learning of local consistency between two views of the same image region in the latent feature space, which is propitious to capture structural representations for image segmentation.
Extensive experiments were conducted on four CT datasets, covering 11 organs and two tumors. Our results not only suggest that using the pre-trained PGL to initialize a segmentation network can dramatically improve its performance, but also indicate the superiority of the proposed PGL model over BYOL, which uses a global consistency constraint.

{\small
\bibliographystyle{ieee_fullname}
\bibliography{egbib}
}

\newpage 
\appendix
\renewcommand{\appendixname}{Appendix~\Alph{section}}

\section{Appendix}
In this part, we provide more technical details and experimental results to supplement the main manuscript. 
We first introduce the detailed network architecture and loss function used in the downstream segmentation tasks (Section~\ref{Sec.Details_Downstream}). 
We then provide more detailed results for `Robustness to less annotations' (Section~\ref{Sec.Robustness}).
Finally, we visualize segmentation results for a qualitative comparison (Section~\ref{Sec.visual}).

\begin{figure*}[!h]
	\begin{center}
		{\includegraphics[width=0.85\linewidth]{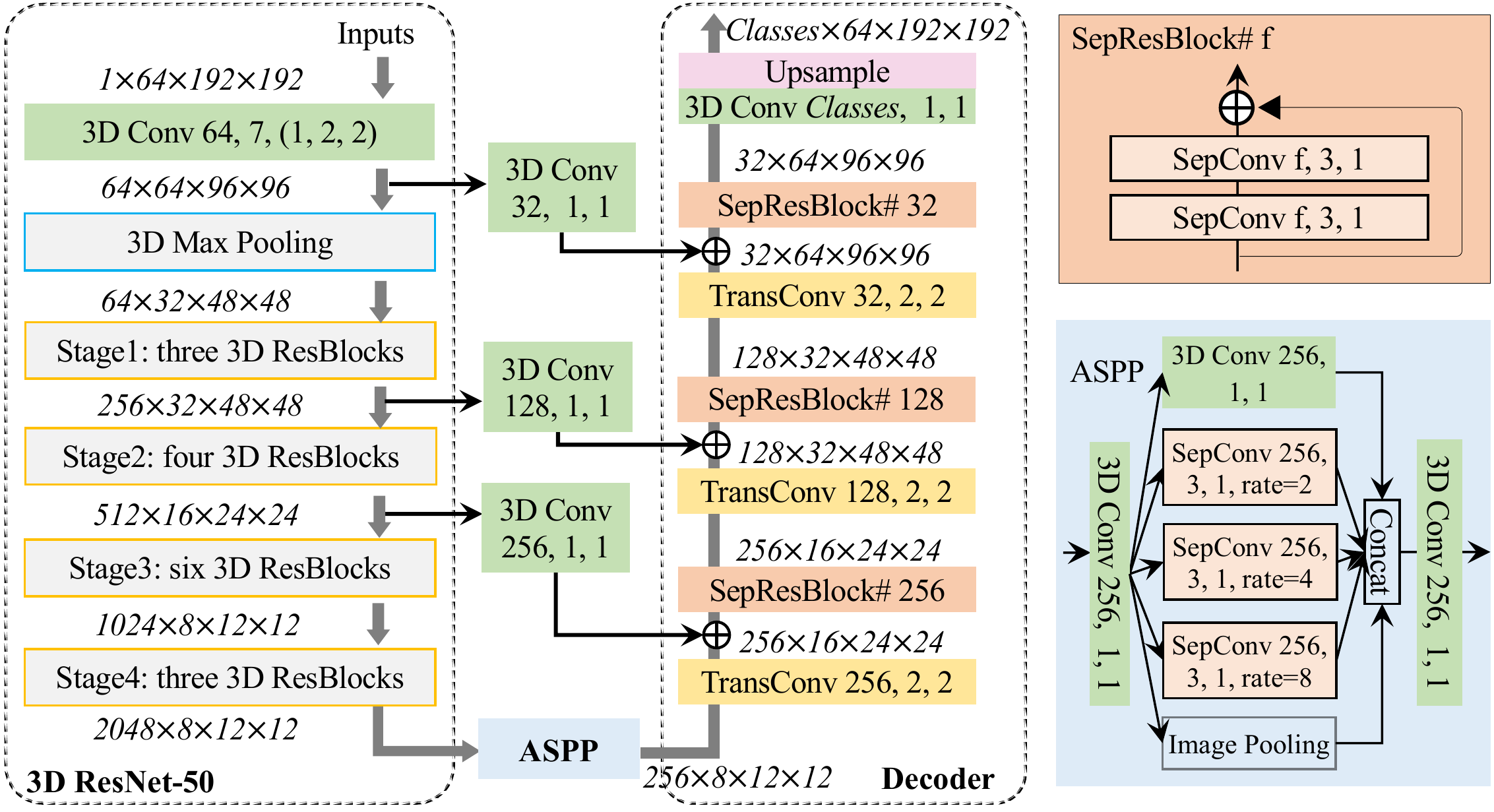}}
	\end{center}
	\vspace{-0.2cm}
	\caption{Detailed architecture of segmentation network: A 3D ResNet-50~\cite{ResNet} encoder, an ASPP module, a decoder, and three skip connections between encoder and decoder. Green `Conv': conv-bn-relu block that contains a 3D convolutional layer followed by a batch normalization (BN) layer and ReLU activation; Yellow `TransConv': 3D transposed convolutional layer; Orange `SepConv': 3D separable convolutional layer, where `rate' denotes the dilation rate; Orange `SepResConv': residual block with two $3\times3$ SepConv layers. Note that the numbers in each block $/$ layer indicate the number of filters, kernel size, and stride, respectively. 
	}	
	\label{fig:fig1}
\end{figure*}

\section{Details on Downstream Tasks}
\label{Sec.Details_Downstream}
\subsection{Architecture of segmentation network}
Figure~\ref{fig:fig1} shows the architecture of the segmentation network designed for downstream segmentation tasks. It consists of a 3D ResNet-50~\cite{ResNet} encoder, an atrous spatial pyramid pooling (ASPP) module~\cite{ASPP}, a decoder, and three skip connections between the encoder and decoder.
The ASPP module has three $1\times1$ convolutional (Conv) layers, an image pooling layer, and three SepConv layers with dilation rates of 2, 4, and 8, respectively.
The decoder contains four upsampling modules.
Each of first three modules has a TransConv layer followed by a SepResConv block, and a pixel-wise summation with the corresponding feature maps from the encoder and the TransConv layer. The last module comprises of a $1\times1$ Conv layer that maps each 32-channel feature maps to the desired number of classes, followed by a Upsampling layer. 
In our experiments, the 3D ResNet-50 encoder was pre-trained by our PGL model, and other parts of the segmentation network are randomly initialized. 

\subsection{loss function}
We jointly use the Dice loss and cross-entropy loss as the loss function of the segmentation network, which is popular in many medical image segmentation applications and has achieved prominent success~\cite{Loss_Zhang, Unet++}. 
On the Liver, Spleen and KiTS datasets, since we only segment one type of organs (and related tumors), we treat the organ or tumor segmentation as binary segmentation tasks. 
The total loss function is formulated as 
\begin{equation}
\begin{aligned}
\small
\mathcal{L} = \left [1 - \frac{2 \sum\tilde{Y} Y}{ \sum(\tilde{Y} + Y + \varepsilon)}\right ]- 
\\
\mathbb{E}\left [Y \log \tilde{Y} + (1-Y)\log(1-\tilde{Y})\right ]
\end{aligned}
\end{equation}
where the first item is the Dice loss, the second item is the binary cross-entropy loss, the prediction and ground truth are denoted by $\tilde{Y}$ and $Y$, respectively, $\mathbb{E}$ is the expectation operation, and $\varepsilon$ is a smoothing factor. 

On the BCV dataset, we jointly use the Dice loss and multi-class cross-entropy loss to segment 13 organs. The total loss function is formulated as 
\begin{equation}
\begin{aligned}
\small
\mathcal{L}=\frac{1}{C}\sum_{n=1}^{C}\left \{\left [1-\frac{2\sum \tilde{Y}^{n}Y^{n}}
{\sum(\tilde{Y}^{n}+Y^{n}+ \varepsilon)}  \right ]-\mathbb{E}\left [Y^{n}log\tilde{Y}^{n} \right ] \right \}
\end{aligned}
\end{equation}
where $C$ is the number of classes.

\section{Detailed Results for Section `Robustness to less annotations'}
\label{Sec.Robustness}
In this section, we provide the comparative experimental results obtained by using the segmentation network to segment 13 organs on the BCV dataset, when using 20\%, 50\% and 100\% of labeled training scans.
The segmentation network was initialized in three ways: (1) random initialization, (2) using the pre-trained BYOL~\cite{BYOL} for intimation, or (3) using the pre-trained PGL for intimation.
The results were displayed in Figure~\ref{fig:fig2}. It reveal that, no matter how many labeled scans were used for fine-tuning, initializing the segmentation network with pre-trained PGL not only steadily outperforms random initialization in the segmentation of all organs, but also beats the BYOL-based initialization in the segmentation of 9 organs.
Moreover, our PGL-based pre-training of the segmentation network results in larger performance gains over random initialization on most segmentation tasks when the labeled training dataset is small.

\begin{figure*}[!h]
	\begin{center}
		{\includegraphics[width=1\linewidth]{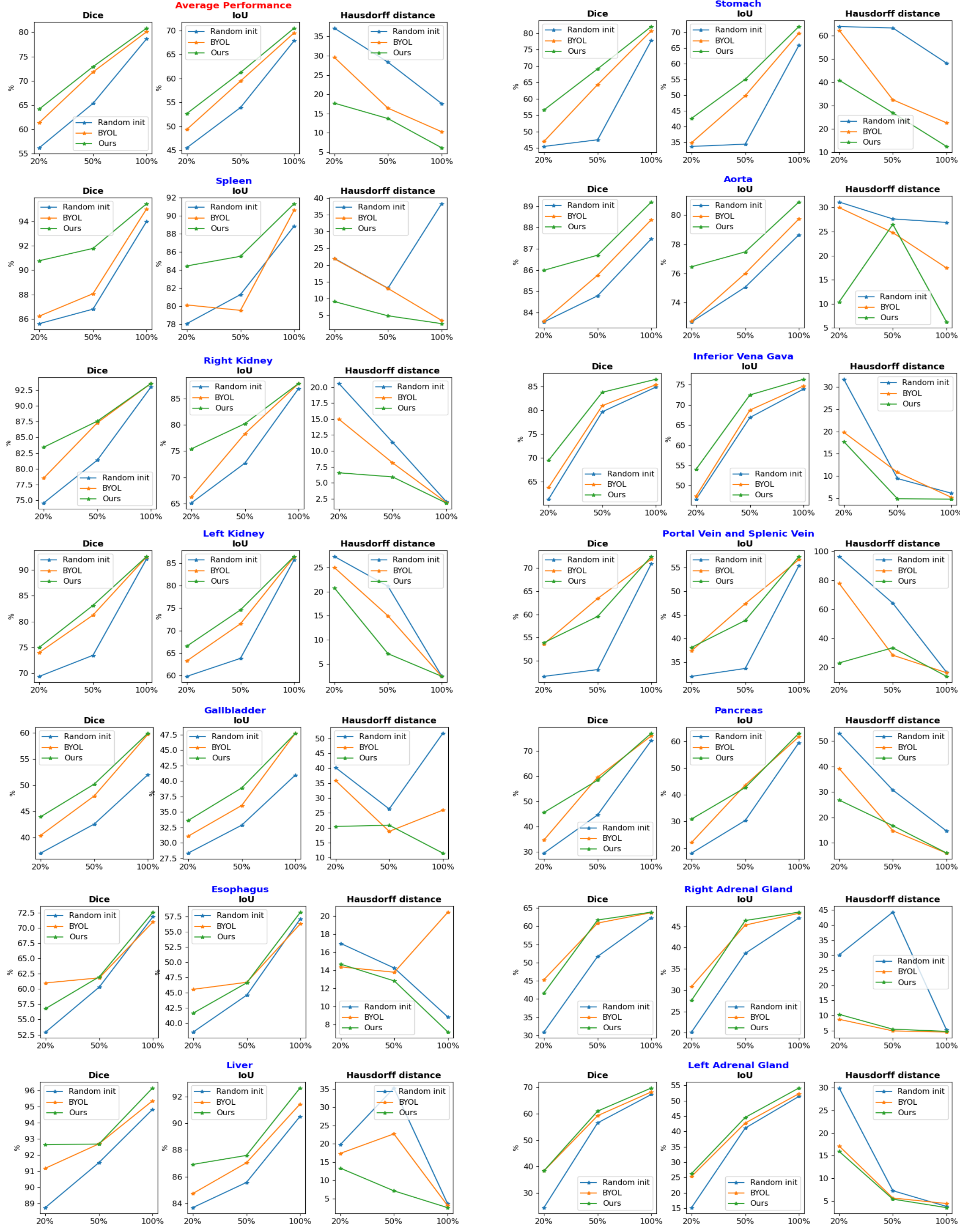}}
	\end{center}
	\vspace{-0.2cm}
	\caption{Performance of segmentation network initialized randomly, using the pre-trained BYOL, or using the pre-trained PGL in segmentation of 13 organs on BCV dataset, when using 20\%, 50\%, or 100\% of labeled downstream training scans.
	}	
	\label{fig:fig2}
\end{figure*}

\section{Visualization of Segmentation Results}
\label{Sec.visual}
The segmentation results produced by the segmentation network, which was initialized (1) randomly, (2) using the pre-trained BYOL~\cite{BYOL}, or (3) using the pre-trained PGL, were visually compared in Figure~\ref{fig:fig3}.
It confirms that, comparing to random initialization and BYOL-based pre-training, our PGL-based pre-training of the segmentation network enables the network to produce the segmentation results, which are more similar to the ground truth. Therefore, our PGL model that poses a local consistency constraint on the aligned features is more beneficial for downstream segmentation tasks than BYOL, which uses a global consistency constraint. 

\begin{figure*}[!h]
	\begin{center}
		{\includegraphics[width=1\linewidth]{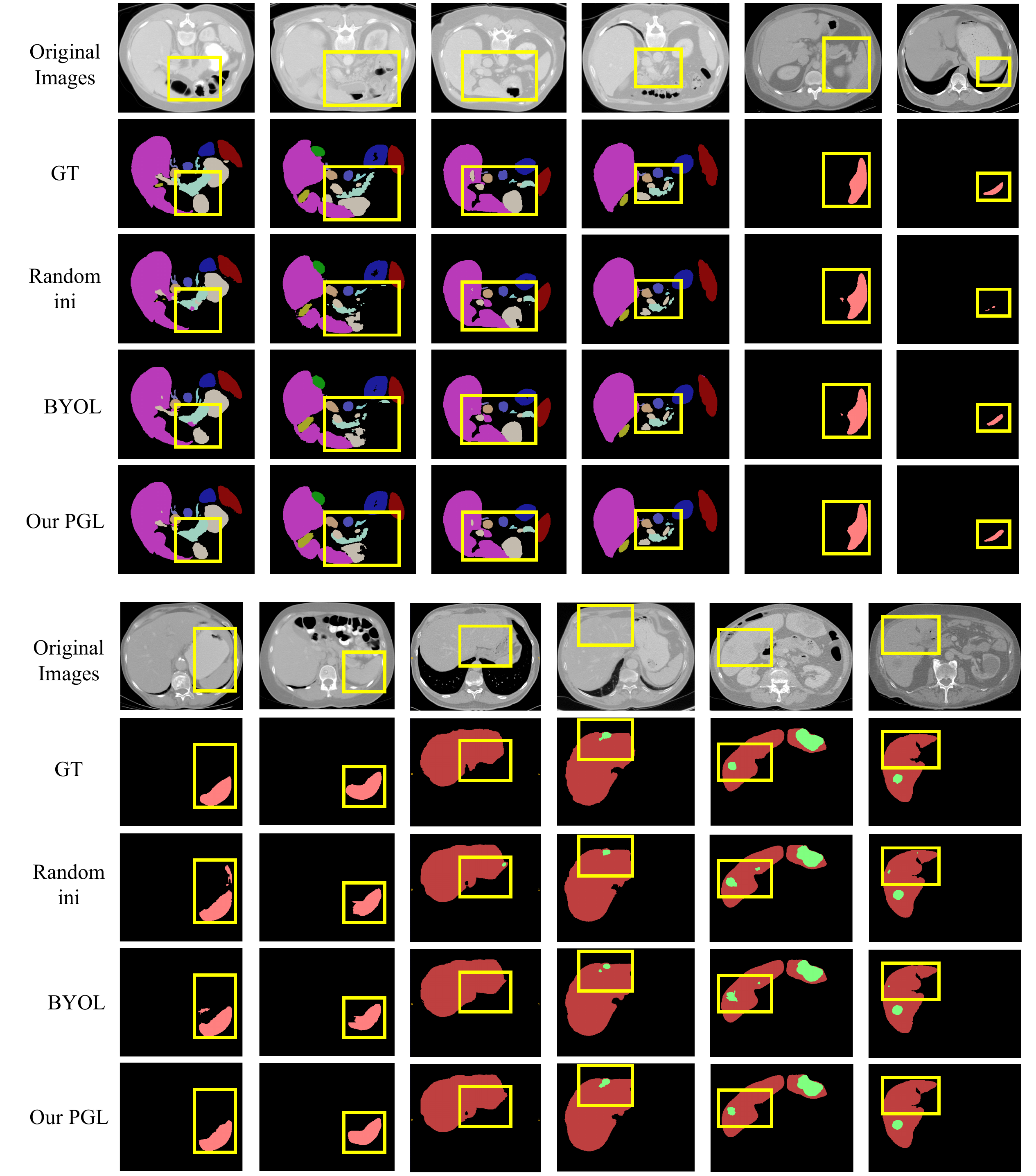}}
	\end{center}
	\vspace{-0.2cm}
	\caption{
	Visualization of segmentation results of 12 cases. The regions in yellow rectangles indicate our superiority. Our PGL-based pre-training results in more accurate results than random initialization and BYOL-based pre-training. Each type of organs and tumors are denoted by a unique color.
	}	
	\label{fig:fig3}
\end{figure*}


\end{document}